# Deliberation Scheduling for Time-Critical Sequential Decision Making

Thomas Dean, Leslie Pack Kaelbling, Jak Kirman, Ann Nicholson
Department of Computer Science
Brown University, Providence, RI 02912

## Abstract

We describe a method for time-critical decision making involving sequential tasks and stochastic processes. The method employs several iterative refinement routines for solving different aspects of the decision making problem. This paper concentrates on the meta-level control problem of *deliberation scheduling*, allocating computational resources to these routines. We provide different models corresponding to optimization problems that capture the different circumstances and computational strategies for decision making under time constraints. We consider *precursor* models in which all decision making is performed prior to execution and *recurrent* models in which decision making is performed in parallel with execution, accounting for the states observed during execution and anticipating future states. We describe algorithms for precursor and recurrent models and provide the results of our empirical investigations to date.

## 1 Introduction

We are interested in solving sequential decision making problems given a model of the underlying dynamical system specified as a stochastic automaton (*i.e.*, a set of states, actions, and a transition matrix which we assume is sparse). In the following, we refer to the specified automaton as the *system* automaton. Our approach builds on the theoretical work in operations research and the decision sciences for posing and solving sequential decision making problems, but it draws its power from the goal-directed perspective of artificial intelligence. Achieving a goal corresponds to performing a sequence of actions in order to reach a state satisfying a given proposition. In general, the shorter the sequence of actions the better. Because the state transitions are governed by a stochastic process, we cannot guarantee the length of a sequence achieving a given goal. Instead, we are interested in minimizing the expected number of actions required to reach the goal.

We represent goals of achievement in terms of an optimal sequential decision making problem in which there is a reward function specially formulated for a particular goal. For the goal of achieving $p$ as quickly as possible, the reward is 0 for all states satisfying $p$ and -1 otherwise. The optimization problem is to find a policy (a mapping from states to actions) maximizing the expected discounted cumulative reward with respect to the underlying stochastic process and the specially formulated reward function. In our formulation, a policy is nothing more than a conditional plan for achieving goals quickly on average.

Instead of generating an optimal policy for the system automaton, which would be impractical for an automaton with a large state space, we formulate a simpler or *restricted* stochastic automaton and then search for an optimal policy in this restricted automaton. At all times, the system maintains a restricted automaton. The restricted automaton and corresponding policy are improved as time permits by successive refinement. This approach was inspired by the work of Drummond and Bresina [Drummond and Bresina, 1990] on anytime synthetic projection.

The state space for the restricted automaton corresponds to a subset of the states of the system automaton (this subset is called the *envelope* of the restricted automaton) and a special state OUT that represents being in some state outside of the envelope. For states in the envelope, the transition function of the restricted automaton is the same as in the system automaton. The pseudo state OUT is a sink (*i.e.*, all actions result in transitions back to OUT) and, for a given action and state in the envelope, the probability of making a transition to OUT is one minus the sum of the probabilities of making a transition to the same or some other state in the envelope.

There are two basic types of operations on the restricted automaton. The first is called *envelope alteration* and serves to increase or decrease the number of states in the restricted automaton. The second is called *policy generation* and determines a policy for the system automaton using the restricted automaton. Note that, while the policy is constructed using the restricted automaton, it is a complete policy and applies to all of the states in the system automaton. For states outside of the envelope, the policy is defined by a set of *reflexes* that implement some default behavior for the agent. In this paper, *deliberation scheduling* refers to the problem of allocating processor time to envelope alteration and policy generation.

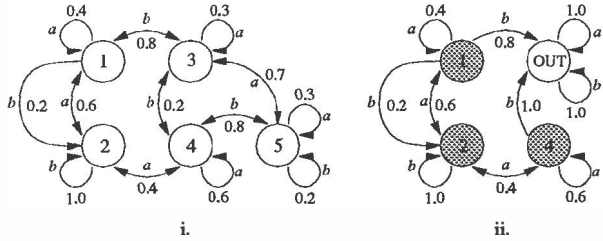

Figure 1: Stochastic process and a restricted version

There are several different methods for envelope alteration. In the first method, we simply search for a (new) path or *trajectory* from the initial state to a state satisfying the goal and add the states traversed in this path to the state space for the restricted automaton. This method need not make use of the current restricted automaton. A second class of methods operates by finding the first state outside the envelope that the agent is most likely to transition to using its current policy, given that it leaves the set of states corresponding the current envelope. There are several variations on this: add the state, add the state and the $n$ next most likely states, add all of the states in a path from the state to a state satisfying the goal, add all of the states in a path from the state to a state back in the current envelope. Finally, there are methods that prune states from the current envelope on the grounds that the agent is unlikely to end up in those states and therefore need not consider them in formulating a policy.

Figure 1.i shows an example system automaton consisting of five states. Suppose that the initial state is 1, and state 4 satisfies the goal. The path $1 \xrightarrow{a} 2 \xrightarrow{a} 4$ goes from the initial state to a state satisfying the goal and the corresponding envelope is $\{1, 2, 4\}$. Figure 1.ii shows the restricted automaton for that envelope. Let $\pi(x)$ be the action specified by the policy $\pi$ to be taken in state $x$; the optimal policy for the restricted automaton shown in Figure 1.ii is defined by $\pi(1) = \pi(2) = \pi(4) = a$ on the states of the envelope and the reflexes by $\pi(\text{OUT}) = b$ (*i.e.*, $\forall x \notin \{1, 2, 4\}, \pi(x) = b$).

All of our current policy generation techniques are based on iterative algorithms such as *value iteration* [Bellman, 1957] and *policy iteration* [Howard, 1960]. In this paper, we use the latter. These techniques can be interrupted at any point to return a policy whose value improves in expectation on each iteration. Each iteration of policy iteration takes $0(|E|^3)$ where $E$ is the envelope or set of states for the restricted automaton. The total number of iterations until no further improvement is possible varies but is guaranteed to be polynomial in $|E|$. This paper is primarily concerned with how to allocate computational resources to envelope alteration and policy generation. In the following, we consider several different models.

In the simpler models called *precursor-deliberation* models, we assume that the agent has one opportunity to generate a policy and that, having generated a policy, the agent must use that policy thereafter. Precursor-deliberation models include

1. a deadline is given in advance, specifying when to stop deliberating and start acting according to the generated policy
2. the agent is given an unlimited amount of time to respond, with a linear cost of delay

There are also more complicated precursor-deliberation models, which we do not address in this paper, such as the following two models, in which a trigger event occurs, indicating that the agent must begin following its policy immediately with no further refinement.

3. the trigger event can occur at any time in a fixed interval with a uniform distribution
4. the trigger event is governed by a more complicated distribution, *e.g.*, a normal distribution centered on an expected time

In more complicated models, called *recurrent-deliberation* models, we assume that the agent periodically replans. Recurrent-deliberation models include

1. the agent performs further envelope alteration and policy generation if and only if it 'falls out' of the envelope defined by the current restricted automaton
2. the agent performs further envelope alteration and policy generation periodically, tailoring the restricted automaton and its corresponding policy to states expected to occur in the near future

The rest of this paper assumes some familiarity with basic methods for sequential decision making in stochastic domains. A companion paper [Dean *et al.*, 1993] provides additional details regarding algorithms for precursor-deliberation models. In this paper, we dispense with the mathematical preliminaries, and concentrate on conveying basic ideas and empirical results. A complete description of our approach including relevant background material is available in a forthcoming technical report.

## 2 Deliberation Scheduling

In the previous section, we sketched an algorithm that generates policies. Each policy $\pi$ has some value with respect to an initial state $x_0$; this value is denoted $V_\pi(x_0)$ and corresponds to the expected cumulative reward that results from executing the policy starting in $x_0$. Given a stochastic process and reward function, $V_\pi(x_0)$ is well defined for any policy $\pi$ and state $x_0$. We are assuming that, in time critical applications, it is impractical to compute $V_\pi(x_0)$ for a given policy and initial state and, more importantly, that it is impractical to compute the optimal policy for the entire system automaton.

In order to control complexity, in generating a policy, our algorithm considers only a subset of the state space of the stochastic process. The algorithm starts with an initial policy and a restricted state space (or *envelope*), extends that envelope, and then computes a new policy. We would like it to be the case that the new policy $\pi'$ is an improvement over (or at the very least no worse than) the old policy $\pi$ in the sense that $V_{\pi'}(x_0) - V_\pi(x_0) \geq 0$.

In general, however, we cannot guarantee that the policy will improve without extending the state space to be the entire space of the system automaton, which results in computational problems. The best that we can hope for is that the algorithm improves in *expectation*. Suppose that the initial envelope is just the initial state and the initial policy is determined entirely by the reflexes. The difference $V_{\pi'}(x_0) - V_\pi(x_0)$ is a random variable, where $\pi$ is the reflex policy and $\pi'$ is the computed policy. We would like it to be the case that $E[V_{\pi'}(x_0) - V_\pi(x_0)] > 0$, where the expectation is taken over start states and goals drawn from some fixed distribution. Although it is possible to construct system automata for which even this improvement in expectation is impossible, we believe most moderately benign navigational environments, for instance, are well-behaved in this respect.

Our algorithm computes its own estimate of the value of policies by using a smaller and computationally more tractable stochastic process. Ideally, we would like to show that there is a strong correlation between the estimate that our algorithm uses and the value of the policy as defined above with respect to the complete stochastic process, but for the time being we show empirically that our algorithm provides policies whose values increase over time.

Our basic algorithm consists of two stages: envelope alteration ($EA$) followed by policy generation ($PG$). The algorithm takes as input an envelope and a policy and generates as output a new envelope and policy. We also assume that the algorithm has access to the state transition matrix for the stochastic process. In general, we assume that the algorithm is applied in the manner of iterative refinement, with more than one invocation of the algorithm. We will also treat envelope alteration and policy generation as separate, so we cast the overall process of policy formation in terms of some number of rounds of envelope alteration followed by policy generation, resulting in a sequence of policies. Figure 2 depicts a sequence of automata generated by iterative refinement along with the associated paths through state space traversed in extending the envelope.

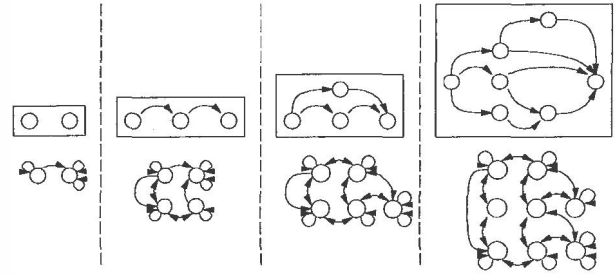

Figure 2: Sequence of restricted automata and associated paths through state space

Envelope alteration can be further classified in terms of three basic operations on the envelope: *trajectory planning*, *envelope extension*, and *envelope pruning*. Trajectory planning consists of searching for some path from an initial state to a state satisfying the goal. Envelope extension consists of adding states to the envelope. Envelope pruning involves removing states from the envelope and is generally used only in recurrent-deliberation models.

Let $\pi_i$ represent the policy after the $i$th round and let $t_{EA_i}$ be the time spent in the $i$th round of envelope alteration. We say that policy generation is *inflexible* if the $i$th round of policy generation is always *run to completion* on $|E_i|$. Policy generation is itself an iterative algorithm that improves an initial policy by estimating the value of policies with respect to the restricted stochastic process mentioned earlier. When run to completion, policy generation continues to iterate until it finds a policy that it cannot improve with respect to its estimate of value. The time spent on the $i$th round of policy generation $t_{PG_i}$ depends on the size of the state space $|E_i|$.

In the following, we present a number of decision models. Note that for each instance of the problems that we consider, there is a large number of possible decision models. Our selection of which decision models to investigate is guided by our interest in providing some insight into the problems of time-critical decision making and our anticipation of the combinatorial problems involved in deliberation scheduling.

## 3  Precursor Deliberation

In this section we consider the first precursor-deliberation model, in which there is a fixed deadline known in advance. It is straightforward to extend this to model 2, where the agent is given an unlimited response time with a linear cost of delay; models 3 and 4 are more complicated and and are not considered in this paper.

### 3.1  The Model

Let $t_{TOT}$ be the total amount of time from the current time until the deadline. If there are $k$ rounds of envelope alteration and policy generation, then we have

$$t_{EA_1} + t_{PG_1} + \cdots + t_{EA_k} + t_{PG_k} = t_{TOT}.$$

**Case I: Single round; inflexible policy generation**  In the simplest case, policy generation does not inform envelope alteration and so we might as well do all of the envelope alteration before policy generation, and $t_{EA_1} + t_{PG_1} = t_{TOT}$. Here is what we need in order to schedule time for $EA_1$ and $PG_1$:

1. the expected value, taken over randomly-chosen pairs of initial states and goals, of the improvement of the value of the initial state, given a fixed amount of time allocated to envelope alteration, $E[V_{\pi_1}(x_0) - V_{\pi_0}(x_0)|t_{EA_1}]$;

2. the expected size of the envelope given the time allocated to the first round of envelope alteration, $E[|E_1| \,| t_{EA_1}]$; and

3. the expected time required for policy generation, given the size of the envelope after the first round of envelope alteration, $E[t_{PG_1}||E_1|]$.

   Note that, because policy generation is itself an iterative refinement algorithm, we can interrupt it at any point and obtain a policy, for instance, when policy generation takes longer than predicted by the above expectation.

Each of (1), (2) and (3) can be determined empirically, and, at least in principle, the optimal allocation to envelope alteration and policy generation can be determined.

**Case II: Multiple rounds; inflexible policy generation**  Assume that policy generation can profitably inform envelope alteration, *i.e.*, the policy after round $i$ provides guidance in extending the environment during round $i+1$. In this case, we also have $k$ rounds and $t_{EA_1} + t_{PG_1} + \cdots + t_{EA_k} + t_{PG_k} = t_{TOT}$.

Informally, let the *fringe states* for a given envelope and policy correspond to those states outside the envelope that can be reached with some probability greater than zero in a single step by following the policy starting from some state within the envelope. Let the most likely *falling-out* state with respect to a given envelope and policy correspond to that fringe state that is most likely to be reached by following the policy starting in the initial state. We might consider a very simple method of envelope alteration in which we just add the most likely falling-out state and then the next most likely and so on. Suppose that adding each additional state takes a fixed amount of time. Let

$$E[V_{\pi_i}(x_0) - V_{\pi_{i-1}}(x_0)| |E_{i-1}| = m, |E_i| = m+n]$$

denote the expected improvement after the $i$th round of envelope alteration and policy generation given that there are $n$ states added to the $m$ states already in the envelope after round $i-1$.

Again, the expectations described above can be obtained empirically. Coupled with the sort of expectations described for Case I (*e.g.*, $E[t_{PG_i}||E_i|]$), one could (in principle) determine the optimal number of rounds $k$ and the allocation to $t_{EA_i}$ and $t_{PG_i}$ for $1 \leq j \leq k$. In practice, we use slightly different statistics and heuristic methods for deliberation scheduling to avoid the combinatorics.

**Case III: Single round: flexible policy generation**  Actually, this case is simpler in concept than Case I, assuming that we can compile the following statistics.

$$E[V_{\pi_1}(x_0) - V_{\pi_0}(x_0)|t_{EA_1}, t_{PG_1}]$$

**Case IV: Multiple round: flexible policy generation**  Again, with additional statistics, *e.g.*,

$$E[V_{\pi_i}(x_0) - V_{\pi_{i-1}}(x_0)| |E_{i-1}| = m, |E_i| = m+n, t_{PG_{i-1}}],$$

this case is not much more difficult than Case II.

### 3.2  Algorithms and Experimental Results

Our initial experiments are based on stochastic automata with up to several thousand states; automata were chosen to be small enough that we can still compute the optimal policy using exact techniques for comparison, but large enough to exercise our approach. The domain, mobile-robot path planning, was chosen so that it would be easy to understand the policies generated by our algorithms. For the experiments reported here, there were 166 locations that the robot might find itself in and four possible orientations resulting in 664 states. These locations are arranged on a grid representing the layout of the fourth floor of the Brown University Computer Science department. The robot is given a task to navigate from some starting location to some target location. The robot has five actions: stay, go forward, turn right, turn left, and turn about. The stay action succeeds with probability one, the other actions succeed with probability 0.8, except in the case of *sinks* corresponding to locations that are difficult or impossible to get out of. In the mobile-robot domain, a sink might correspond to a stairwell that the robot could fall into. The reward function for the sequential decision problem associated with a given initial and target location assigns 0 to the four states corresponding to the target location and $-1$ to all other states.

We gathered a variety of statistics on how extending the envelope increases value. The statistics that proved most useful corresponded to the expected improvement in value for different numbers of states added to the envelope. Instead of conditioning just on the size of the envelope prior to alteration we found it necessary to condition on both the size of the envelope and the estimated value of the current policy (*i.e.*, the value of the optimal policy computed by policy iteration on the restricted automaton). At run time, we use the size of the automaton and the estimated value of the current policy to index into a table of *performance profiles* giving expected improvement as a function of number of states added to the envelope. Figure 3 depicts some representative functions for different ranges of the value of the current policy.

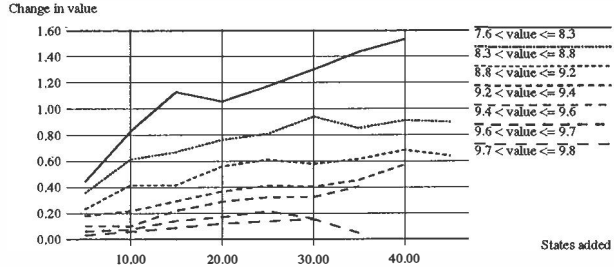

Figure 3: Expected improvement as a function of the number of states $n$ added to initial envelope of size $m$

In general, computing the optimal deliberation schedule for the multiple-round precursor-deliberation models described above is computationally complex. We have experimented with a number of simple, greedy and myopic scheduling strategies; we report on one such strategy here.

Using the mobile-robot domain, we generated 380,000 data points to compute statistics of the sort shown in Figure 3 plus estimates of the time required for one round of envelope alteration followed by policy generation given the size of the envelope, the number of states added, and value of the current policy. We use the following simple greedy strategy for choosing the number of states to add to the envelope on each round. For each round of envelope alteration followed by policy generation, we use the statistics to determine the number of states which, added to the envelope, maximizes the ratio of performance improvement to the time required for computation. Figure 4 compares the greedy algorithm with the standard (inflexible) policy iteration on the complete automaton and with an interruptable (flexible) version of policy iteration on the complete automaton. The data for Figure 4 was determined from one representative run of the three algorithms on a particular initial state and goal. In another paper [Dean et al., 1993] we present results for the average improvement of the start state under the policy available at time $t$ as a function of time.

## 4  Recurrent Deliberation

### 4.1  The Model

In recurrent-deliberation models, the agent has to repeatedly decide how to allocate time to deliberation, taking into account new information obtained during execution. In this section, we consider a particular model for recurrent deliberation in which the agent allocates time to deliberation only at prescribed times. We assume that the agent has separate deliberation and execution modules that run in parallel and communicate by message passing; the deliberation module sends policies to the execution module and the execution module sends observed states to the deliberation module. We also assume that the agent correctly identifies its current state; in the extended version of this paper, we consider the case in which there is uncertainty in observation.

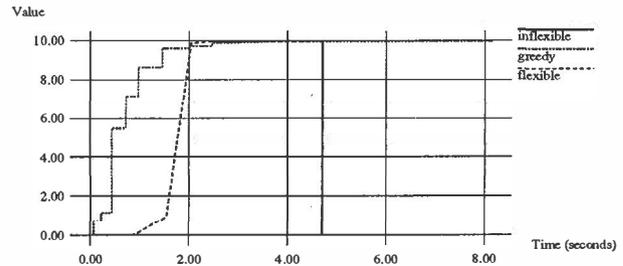

Figure 4: Comparison of the greedy algorithm with standard (inflexible) policy iteration and interruptable (flexible) policy iteration

We call the model considered in this section the *discrete, weakly-coupled, recurrent deliberation* model. It is discrete because each tick of the clock corresponds to exactly one state transition; recurrent because the execution module gets a new policy from the deliberation module periodically; weakly coupled in that the two modules communicate by having the execution module send the deliberation module the current state and the deliberation module send the execution module the latest policy. In this section, we consider the case in which communication between the two modules occurs exactly once every $n$ ticks; at times $n, 2n, 3n, \ldots$, the deliberation module sends off the policy generated in the last $n$ ticks, receives the current state from the execution module, and begins deliberating on the next policy. In the next section, we present an algorithm for the case where the interval between communications is allowed to vary.

In the recurrent models, it is often necessary to remove states from the envelope in order to lower the computational costs of generating policies from the restricted automata. For instance, in the mobile-robot domain, it may be appropriate to remove states corresponding to portions of a path the robot has already traversed if there is little chance of returning to those states. In general, there are many more possible strategies for deploying envelope alteration and policy generation in recurrent models than in the case of precursor models. Figure 5 shows a typical sequence of changes to the envelope corresponding to the state space for the restricted automaton. The current state is indicated

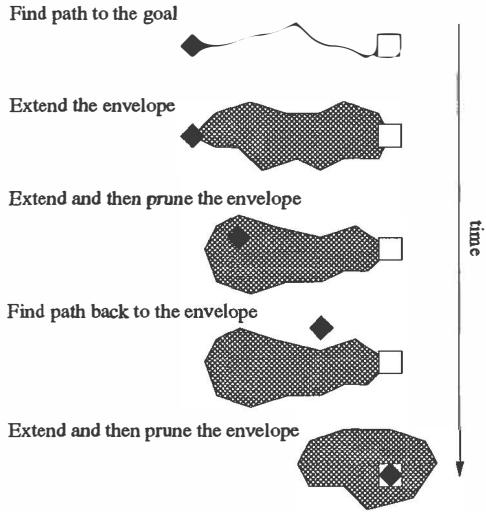

Figure 5: Typical sequence of changes to the envelope

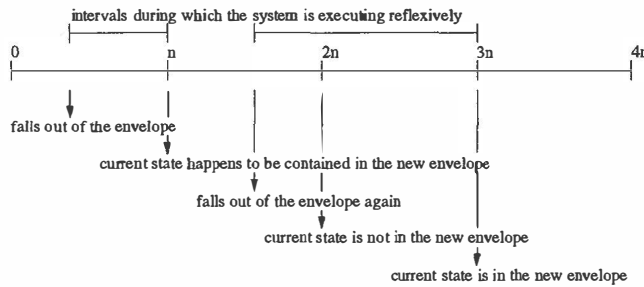

Figure 6: Recurrent-deliberation

by ◆ and the goal state is indicated by □.

To cope with the attendant combinatorics, we raise the level of abstraction and assume that we are given a small set of strategies that have been determined empirically to improve policies significantly in various circumstances. Each strategy corresponds to some fixed schedule for allocating processor time to envelope alteration and policy generation routines. Strategies would be tuned to a particular $n$-tick deliberation cycle. One strategy might be to use a particular pruning algorithm to remove a specified number of states and then use whatever remains of the $n$ ticks to generate a new policy. In this regime, deliberation scheduling consists of choosing which strategy to use at the beginning of each $n$-tick interval. In this section, we ignore the time spent in deliberation scheduling; in the next section, we will arrange it so that the time spent in deliberation scheduling is negligible.

Before we get into the details of our decision model, consider some complications that arise in recurrent-deliberation problems. At any given moment, the agent is executing a policy, call it $\pi$, defined on the current envelope and augmented with a set of reflexes for states falling outside the envelope. The agent begins executing $\pi$ in state $x$. At the end of the current $n$-tick interval, the execution module is given a new policy $\pi'$, and the deliberation module is given the current state $x'$. It is possible that $x'$ is not included in the envelope for $\pi'$; if the reflexes do not drive the robot inside the envelope then the agent's behavior throughout the next $n$-tick interval will be determined entirely by the reflexes. Figure 6 shows a possible run depicting intervals in which the system is executing reflexively and intervals in which it is using the current policy; for this example, we assume reflexes that enable an agent to remain in the same state indefinitely.

Let $\delta_n(x, \pi, x')$ be the probability of ending up in $x'$ starting from $x$ and following $\pi$ for $n$ steps. Suppose that we are given a set of strategies $\{F_1, F_2, \ldots\}$. As is usual in such combinatorial problems with indefinite horizons, we adopt a myopic decision model. In particular, we assume that, at the beginning of each $n$-tick interval, we are planning to follow the current policy $\pi$ for $n$ steps, follow the policy $F(\pi)$ generated by some strategy $F$ attempting to improve on $\pi$ for the next $n$ steps, and thereafter follow the optimal policy $\pi^*$. If we assume that it is impossible to get to a goal state in the next $2n$ steps, the expected value of using strategy $F$ is given by

$$\sum_{i=0}^{2n-1} \gamma^i + \gamma^{2n} \sum_{x' \in X} \delta_n(x, \pi, x') \left[ \sum_{x'' \in X} \delta_n(x', F(\pi), x'') V_{\pi^*}(x'') \right],$$

where $0 <= \gamma < 1$ is a discounting factor, controlling the degree of influence of future results on the current decision.

Extending the above model to account for the possibility of getting to the goal state in the next $2n$ steps is straightforward; computing a good estimate of $V_{\pi^*}$ is not, however. We might use the value of some policy other than $\pi^*$, but then we risk choosing strategies that are optimized to support a particular suboptimal policy when in fact the agent should be able to do much better. In general, it is difficult to estimate the value of prospects beyond any given limited horizon for sequential decision problems of indefinite duration. In the next section, we consider one possible practical expedient that appears to have heuristic merit.

### 4.2 Algorithms and Experimental Results

In this section, we present a method for solving recurrent-deliberation problems of indefinite duration using statistical estimates of the value of a variety of deliberation strategies. We deviate from the decision model described in the previous section in one additional important way; we allow variable-length intervals for deliberation. Although fixed-length facilitate exposition, it is much easier to collect useful statistical estimates of the utility of deliberation strategies if the deliberation interval is allowed to vary.

For the remainder of this section, a deliberation strategy is just a particular sequence of invocations of envelope alteration and policy generation routines. Delib-eration strategies are parameterized according to attributes of the policy such as the estimated value of policies and the size of the envelopes. The function $EIV(F, V_\pi, |E_\pi|)$ provides an estimate of the expected improvement from using the strategy $F$ assuming that the estimated value of the current policy and the size of the corresponding envelope fall within the specified ranges. This function is implemented as a table in which each entry is indexed by a strategy $F$ and a set of ranges, e.g., $\{[\min V_\pi, \max V_\pi], [\min |E_\pi|, \max |E_\pi|]\}$.

We determine the $EIV$ function off line by gathering statistics for $F$ running on a wide variety of policies. The ranges are established so that, for values within the specified ranges the expected improvements have low variance. At run time, the deliberation scheduler computes an estimate of the current policy $\hat{V}_\pi$, determines the size $|E_\pi|$ of the corresponding envelope and chooses the strategy $F$ maximizing $EIV(F, \hat{V}_\pi, |E_\pi|)$.

To build a table of estimates of function $EIV$ off line, we begin by gathering data on the performance of strategies ranging over possible initial states, goals, and policies. For a particular strategy $F$, initial state $x$, and policy $\pi$, we run $F$ on $\pi$, determine the elapsed number of steps $k$, and compute estimated improvement in value,

$$\left[-\sum_{i=0}^{k-1} \gamma^i + \gamma^k \sum_{x' \in X} \delta_k(x, \pi, x') \hat{V}_{F(\pi)}(x')\right] - \hat{V}_\pi(x),$$

where the first term corresponds to the value of using $\pi$ for the first $k$ steps and $F(\pi)$ there after and the second term corresponds to the case in which we do no deliberation whatsoever and use $\pi$ forever. As in the model described in the previous section, we assume that the goal cannot be reached in the next $k$ steps; again it is simple to extend the analysis to the case in which the goal may be reached in less than $k$ steps. Given data of the sort described above, we build the table for $EIV(F, V_\pi, |E_\pi|)$ by appropriately dividing the data into subsets with low variance.

One unresolved problem with this approach is exactly how we are to compute $\hat{V}_\pi(x)$. Recall that $\pi$ is only a partial policy defined on a subset of $X$ augmented with a set of reflexes to handle states outside the current envelope. In estimating the value of a policy, we are really interested in estimating the value of the augmented partial policy. If the reflexes kept the agent in the same place indefinitely, then as long as there was some nonzero probability of falling out of the envelope with a given policy starting in a given state the actual value of the policy in that state would be $-1/(1-\gamma)$. Of course, this is an extremely pessimistic estimate for the long term value of a particular policy since in the recurrent model the agent will periodically compute a new policy based on where it is in the state space. The problem is that we cannot directly account for these subsequent policies without extending the horizon of the myopic decision model and absorbing the associated computational costs in offline data gathering and online deliberation scheduling.

To avoid complicating the online decision making, we have adopted the following expedient which allows us to keep our one-step-lookahead model. We modify the transition probabilities for the restricted automaton so that there is always a non-zero probability of getting back into the envelope having fallen out of it. Exactly what this probability should be is somewhat complicated. The particular value chosen will determine just how concerned the agent will be with the prospect of falling out of the envelope. In fact, the value is dependent on the actual strategies chosen by deliberation scheduling which, in our particular case, depends on $EIV$ and this value of falling back in. We might possibly resolve the circularity by solving a large and very complicated set of simultaneous equations; instead, we have found that in practice it is not difficult to find a value that works reasonably well.

The experimental results for the recurrent model were obtained on the mobile-robot domain with 1422 possible locations and hence 5688 states. The actions available to the agent were the same as those used to obtain the precursor-model results. The transition probabilities were also the same, except that the domain no longer contained sinks.

We used a set of 24 hand-crafted strategies, which were combinations of envelope **optimization** (O) and the following types of envelope alteration;

1. **findpath** (FP): if the agent's current state $x_{cur}$ is not in the envelope, find a path from $x_{cur}$ to a goal state, and add this path to the envelope
2. **robustify** (R[N]): we used the following heuristic to extend the envelope: find the $N$ most likely fringe states that the agent would fall out of the envelope into, and add them to the envelope
3. **prune** (P[N]): of the states that have a worse value than the current state, remove the $N$ least likely to be reached using the current policy.

Each of the strategies used began with **findpath** and ended with **optimization**. Between the first and last phases, robustification, pruning and optimization were used in different combinations, with the number of states to be added or deleted $\in \{10, 20, 50, 100\}$; examples of the strategies we used are {FP R[10] O}, {FP P[20] O}, {FP P[20] R[50] O}, {FP R[100] P[50] O}, {FP R[50] O P[50] O}.

We collected statistics over about 4000 runs generating 100,000 data points for strategy execution. The start/goal pairs were chosen uniformly at random and we ran the simulated robot in parallel with the planner until the goal was reached. The planner executed the following loop: choose one of the 24 strategies uniformly at random, execute that strategy, and then pass the new policy to the simulated robot. We found the following conditioning variables to be significant: the envelope size, $|E|$, the value of the current state $V_\pi$, the "fatness" of the envelope (the ratio of envelope size to fringe size), and the Manhattan distance, $M$, between the start and goal locations. We then build a lookup table of expected improvement in value over the time the strategy takes to compute, $\delta V_\pi / k$, as a function of E, $V_\pi$, the fatness, $M$ and the strategy $s$.

To test our algorithm, we took 25 pairs of start and goal states, chosen uniformly at random from pairs of Manhattan distance less than one third of the diameter of the world. For each pair we ran the simulated robot in parallel with the following deliberation mechanisms:

- recurrent-deliberation with strategies chosen using statistical estimates of $EIV$ (LOOKUP)
- dynamic programming policy iteration over the entire domain, with a new policy given to the robot after each iteration (ITER) and only after it has been optimized (WHOLE)

The average number of steps taken by LOOKUP, ITER and WHOLE were 71, 87 and 246 respectively

While the improvement obtained using the recurrent-deliberation algorithm is only small it is statistically significant. These preliminary results were obtained when there were still bugs in the implementation, however, since we have determined that the strategies are in fact being pessimistic, we expect to obtain further performance improvement using LOOKUP. Recall also that we are still working in the comparatively small domain necessary to be able to compute the optimal policy over the whole domain; for larger domains, ITER and WHOLE are computationally infeasible.

## 5   Related Work and Conclusions

Our primary interest is in applying the sequential decision making techniques of Bellman [Bellman, 1957] and Howard [Howard, 1960] in time-critical applications. Our initial motivation for this research arose in attempting to put the anytime synthetic projection work of Drummond and Bresina [Drummond and Bresina, 1990] on more secure theoretical foundations. The approach described in this paper represents a particular instance of time-dependent planning [Dean and Boddy, 1988] and borrows from, among others, Horvitz' [Horvitz, 1988] approach to flexible computation. Hansson and Mayer's BPS (Bayesian Problem Solver) [Hansson and Mayer, 1989] supports general state space search with decision theoretic control of inference; it may be that BPS could be used as the basis for envelope alteration. Boddy [Boddy, 1991] describes solutions to related problems involving dynamic programming. For an overview of resource-bounded decision making methods, see chapter 8 of the text by Dean and Wellman [Dean and Wellman, 1991].

We have presented an approach to coping with uncertainty and time pressure in decision making. The approach lends itself to a variety of online computational strategies, a few of which are described in this paper. Our algorithms exploit both the goal-directed, state-space search methods of artificial intelligence and the dynamic programming, stochastic decision making methods of operations research. Our empirical results demonstrate that it is possible to obtain high performance policies for large stochastic processes in a manner suitable for time critical decision making.

### Acknowledgements

Tom Dean's work was supported in part by a National Science Foundation Presidential Young Investigator Award IRI-8957601, by the Advanced Research Projects Agency of the DoD monitored by the Air Force under Contract No. F30602-91-C-0041, and by the National Science foundation in conjunction with the Advanced Research Projects Agency of the DoD under Contract No. IRI-8905436. Leslie Kaelbling's work was supported in part by a National Science Foundation National Young Investigator Award IRI-9257592 and in part by ONR Contract N00014-91-4052, ARPA Order 8225. Thanks also to Moises Lejter for his input during the development and implementation of the recurrent deliberation model.